\documentclass[10pt,twocolumn,letterpaper]{article}

\usepackage{iccv}
\usepackage{times}
\usepackage{epsfig}
\usepackage{graphicx}
\usepackage{amsmath}
\usepackage{amssymb}
\usepackage{array}
\usepackage{multirow}
\usepackage[utf8]{inputenc}

\usepackage[breaklinks=true,bookmarks=false]{hyperref}

\iccvfinalcopy 

\ificcvfinal\fi

\newcolumntype{L}[1]{>{\raggedright\let\newline\\\arraybackslash\hspace{0pt}}m{#1}}

\begin{document}

\title{Towards A Fairer Landmark Recognition Dataset}

\author{Zu Kim \quad Andr\'{e} Araujo \quad Bingyi Cao \quad Cam Askew \\ Jack Sim \quad Mike Green \quad N'Mah Fodiatu Yilla \quad Tobias Weyand\\
Google Research\\
{\tt\small \{zkim,andrearaujo,bingyi,askewc,jacksim,greenmike,nyilla,weyand\}@google.com}
}

\maketitle
\ificcvfinal\thispagestyle{empty}\fi

\newcommand{\andre} [1]{{\color{magenta}#1}}
\newcommand{\zkim} [1]{{\color{cyan}#1}}

\begin{abstract}
We introduce a new landmark recognition dataset, which is created with a focus on fair worldwide representation.
While previous work proposes to collect as many images as possible from web repositories, we instead argue that such approaches can lead to biased data.
To create a more comprehensive and equitable dataset, we start by defining the fair {\em relevance} of a landmark to the world population.
These relevances are estimated by combining anonymized Google Maps user contribution statistics with the contributors' demographic information.
We present a stratification approach and analysis which leads to a much fairer coverage of the world, compared to existing datasets.
The resulting datasets are used to evaluate computer vision models as part of the the Google Landmark Recognition and Retrieval Challenges 2021.
\end{abstract}

\section{Introduction}

There has been significant progress in machine learning systems over the last decade, due in part to the growing availability of data at scale for model development.
Many of the widely used machine learning datasets have been collected from the internet, leveraging content uploaded by users from around the world.
An important challenge in this context is that the data may not be represented in a fair manner, since the distribution on internet users does not represent well the world population -- due to differences in accessibility to the internet and computing devices \cite{Jo_2020}.
As a result, datasets created in this manner tend to be biased, leading to machine learning models which may not capture information fairly~\cite{mehrabi2019survey}.

\begin{figure}
\centering
\includegraphics[width=8cm]{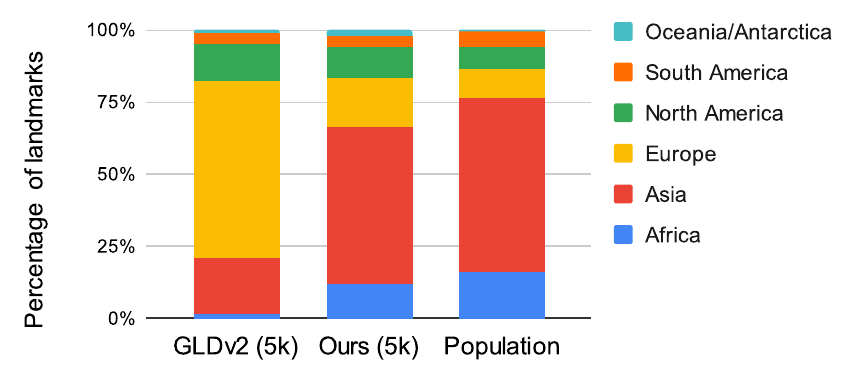}
\caption{Proportion of landmarks from each continent available in different datasets (left, center), compared to the world population distribution (right). The Google Landmarks Dataset v2~\cite{weyand2020gldv2} (GLDv2, left) contains significant biases and does not capture the world population fairly.
Our new dataset (center), constructed using a proposed stratified approach, reflects much better the world population.
For both datasets, we include the top 5000 landmarks in this chart.}
\label{fig:continent-breakdown}
\end{figure}

In this work, we are particularly interested in the computer vision problem of landmark recognition, where the goal is to identify human-made or natural landmarks depicted in images -- for example: famous buildings, mountains, waterfalls, towers, monuments, etc.
Landmark recognition is an active research area, and multiple datasets \cite{weyand2020gldv2,Philbin2007OxfordDataset,Philbin2008ParisDataset} have been created for model training and evaluation.
These projects provide large datasets for academic evaluation, but they failed to consider the representativeness of different data with respect to the inherent bias present in online repositories.
For example, Weyand \etal \cite{weyand2020gldv2} discuss the geographic distribution of their dataset, showing that landmarks in some parts of the world (\eg, Africa, India, China) are significantly under-represented compared with their population.
This illustrates how bias can be directly introduced by simple crowdsourcing of web data.

In this paper, we present an approach to reduce the bias of crowdsourced data. We first define the fair {\em relevance} of a landmark with respect to the world population, and then propose an approach to estimate it -- leveraging anonymized and aggregated user contribution statistics from Google Maps\footnote{\url{https://www.google.com/maps}}. These estimated relevance scores allow us to create a fairer test dataset, which is used to evaluate models as part of the 2021 edition of the Google Landmark Recognition\footnote{\url{https://www.kaggle.com/c/landmark-recognition-2021/}} and Retrieval\footnote{\url{https://www.kaggle.com/c/landmark-retrieval-2021/}} challenges. A baseline evaluation is also performed, serving as a reference for future model development in this area.

\section{Estimating landmark relevances}
\label{sec:approach}

Our approach starts with defining the fair {\em relevance} of a landmark, which depends directly on the specific target application.
In our case, we are interested in the following definitions in particular:

\begin{itemize}
    \item \textbf{Personal Importance / Utility: } “Which places are personally important (or useful)? (e.g., grocery market, school, local social  security office)”
    \item \textbf{Tourism: } “If everyone in the world can go anywhere in the world, what is the place they would like to go/remember?”
\end{itemize}

To answer these questions, we propose to leverage user contributions to web platforms, but crucially debiasing them according to available demographic information.
As previously mentioned, we expect user contributions to internet repositories to be biased because the uploaders are not representative of the general population. 
For example, there is a large variation in internet accessibility and proficiency by place of living, education level, gender, income level, travel ability and age \cite{Jo_2020}.

We leverage a version of stratified sampling~\cite{Neyman_1992_Stratified_Sampling} to compute relevance scores of such crowdsourced data, taking into consideration demographic categories of the uploaders.
Using this approach, we can model the ideal distribution of landmark relevances as:
\begin{equation} \label{eq:raw_fomulation}
\begin{aligned}
P(c) &= \sum_h{P(c|h) P(h)} \\
     &= \sum_{h_1,h_2,\ldots}{P(c|h_1,h_2,\ldots) P(h_1,h_2,\ldots)},
\end{aligned}
\end{equation}
where $P(c)$ is the probability distribution over the landmarks $c$ (representing their relevances); and $h = \{h_1, h_2, \ldots\}$ represents a demographic group (strata), based on specific demographic categories $h_i$.
For example, $h$ may represent Canadian females, where $h_1$ would denote Canadian residents and $h_2$ females.
$P(c|h)$ is the distribution of landmark relevances for contributors who belong to the demographic group $h$.
This distribution can be estimated from concrete contribution statistics from the given demographic group:

\[
P(c|h) = \frac{A_{c,h}}{A_{h}},
\]
Where $A_{c,h}$ is the number of contributions of the demographic group $h$ to the landmark $c$ and $A_{h}$ is the total number of contributions of the demographic group $h$ to all landmarks.

Naive crowdsourcing without bias adjustment essentially assumes that the prior probability of a demographic group, $P(h)$, follows the contributors’ demographic distribution.
Instead, we use the world’s population statistics to estimate $P(h)$, which improves the estimate of $P(c)$ since it avoids the biases of contributors.
Demographic information can be obtained from various sources that provide such statistics, such as \cite{PopulationPyramid, DataCommons}.
For example, a prior for Canadian ($h_1$) females ($h_2$),
\[
P(h_1, h_2) = \frac{N_{h_1,h_2}}{N_W},
\]
where $N_{h_1,h_2}$ is the Canadian female population and $N_W$ is the world population.


\section{A Fairer Landmark Recognition Dataset}

\paragraph{Dataset.}
We leverage Google Maps, which is an online platform with a large number of user contributions: image uploads, ratings, and reviews.
We consider user contribution statistics to this platform as a proxy for landmark relevance.
For data stratification, we leverage three types of demographic categories: gender, age and country of residence.
Note that such demographic information is highly private, so we performed thorough reviews and procedures to make sure that these information cannot be inferred from the result.

\paragraph{Implementation details.}
The exact procedures we used to estimate landmark relevances are more complex than the ones described in Section \ref{sec:approach} -- this is needed in order to handle unavailable statistics (e.g., non-binary gender).
This will be described in more detail in a future publication.

\paragraph{Results.}
Using Google Maps user contribution information and demographic statistics from \cite{PopulationPyramid, DataCommons}, we created relevance scores for 50,000 landmarks.

\begin{figure}
\centering
\includegraphics[width=8cm]{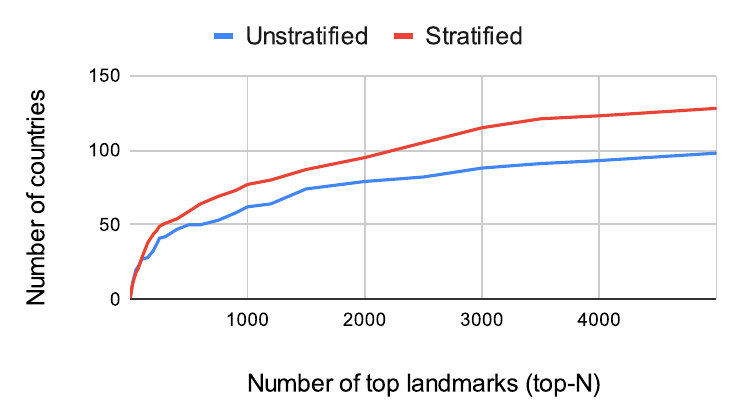}
\caption{Number of countries that host the top-N landmarks. We see that the stratification increases diverse representation.}
\label{fig:country-breakdown}
\end{figure}

Table~\ref{table:top10-stratified} shows the top-10 landmarks based on our relevance scores and Figure~\ref{fig:country-breakdown} shows the number of countries that host the top-N landmarks (for a variety of N). We see that the stratification increases diverse representation of countries that have been under-represented.

In Figure~\ref{fig:continent-breakdown}, we compare the continent breakdown of top-5000 landmarks in our dataset and the Google Landmarks Dataset~\cite{weyand2020gldv2}. We see that our dataset better represents the population of each continent.

\begin{table*}
\begin{center}
\begin{tabular}{|c|c|c|}
\hline
Rank & Top-10 landmarks (Unstratified) & Top-10 landmarks (Stratified) \\
\hline
\hline
1 & Eiffel Tower (France) & Eiffel Tower (France) \\
\hline
2 & Colosseum (Italy) & Masjid al-Haram (Saudi Arabia) \\
\hline
3 & Trevi Fountain (Italy) & Taj Mahal (India) \\
\hline
4 & Taj Mahal (India) & Al Masjid an Nabawi (Saudi Arabia) \\
\hline
5 & Louvre Museum (France) & Louvre Museum (France) \\
\hline
6 & Gateway Of India Mumbai (India) & Gateway Of India Mumbai (India) \\
\hline
7 & Masjid al-Haram (Saudi Arabia) & The Dubai Mall (Arab Emirates) \\
\hline
8 & Central Park (US) & Mysore Palace (India) \\
\hline
9 & Mysore Palace (India) & Giza Necropolis (Egypt) \\
\hline
10 & Walt Disney World\textsuperscript{\textregistered} Resort (US) & Colosseum (Italy) \\
\hline
\end{tabular}
\end{center}
\caption{Top-10 landmarks based on the relevance score.}
\label{table:top10-stratified}
\end{table*}

\paragraph{Datasets for Google Landmarks Challenges.}

We used the relevance scores to create fairer evaluation datasets for the 2021 version of the Google Landmark Recognition and Retrieval challenges.
The following datasets were created by resampling the images in GLDv2 based on the distribution of the relevance scores. Landmarks without corresponding relevance scores (e.g., long-tail landmarks) were still subjects of sampling; we assigned the lowest available relevance score to them.
\begin{itemize}
    \item \textbf{Index dataset (retrieval challenge):} 100,000 images sampled from the GLDv2 training dataset.
    \item \textbf{Public eval dataset (retrieval challenge):} 514 images sampled/downloaded from the GLDv2 training dataset and Wikimedia. This only contains the images of the landmarks that are present in the above index dataset.
    \item \textbf{Private eval dataset (retrieval challenge):} 1028 sampled/downloaded from the GLDv2 training dataset and Wikimedia. This only contains the images of the landmarks that are present in the above index dataset.
    \item \textbf{Index dataset (recognition challenge):} 100,000 images sampled from the GLDv2 training dataset.
    \item \textbf{Public eval dataset (recognition challenge):} 9735 images sampled/downloaded from Wikimedia. This contains images of the landmarks that are present in the above (recognition challenge) index dataset and non-landmark (distractor) images.
    \item \textbf{Private eval dataset (recognition challenge):} 10502 images sampled/downloaded from Wikimedia. This contains images of the landmarks that are present in the above (recognition challenge) index dataset and non-landmark (distractor) images.
\end{itemize}

The baseline evaluation results with the above datasets are shown in Table~\ref{table:evaluation}. We used DELG global embedding~\cite{cao2020delg} with nearest neighbor matching for both recognition and retrieval results.
See Section~\ref{sec:discussion} for a further discussion on the limitations of these datasets.

\begin{table*}
\begin{center}
\begin{tabular}{c|c|c}
\hline
Data & Private & Public \\
\hline
2020 retrieval challenge (mAP@100) & 0.2407 & 0.2258 \\
2021 retrieval challenge (mAP@100) & 0.2223 & 0.2153 \\
2020 recognition challenge ($\mu$AP@100) & 0.2244 & 0.2293 \\
2021 recognition challenge ($\mu$AP@100) & 0.2303 & 0.2376 \\
\hline
\end{tabular}
\end{center}
\caption{Baseline results with the new datasets in comparison with the previous year’s. DELG global embedding~\cite{cao2020delg} was used for the evaluation.}
\label{table:evaluation}
\end{table*}


\section{Discussion}
\label{sec:discussion}

We presented an approach to reduce bias in landmark recognition and applied it to create a new dataset. Our analysis shows that the presented approach effectively increases the diverse representation. 

\paragraph{Limitations.}
There still are limitations in our method:

\begin{itemize}
    \item Some of the potential sources of bias (\eg, internet proficiency, income level, ability to travel) were not addressed or only implicitly addressed through correlated variables (gender, age, and country).
    \item We may still not have enough landmarks in some of the under-represented regions due to a lack of internet access and/or users in the region.
    \item The relevance score is only available for a subset of the Google Landmarks Dataset V2 (GLDv2). Please refer to \cite{weyand2020gldv2} on the limitation of the GLDv2 dataset’s coverage. In addition, a relevance score may not be provided for a highly relevant landmark due to various reasons including the incompatibility between Google Maps and Wikimedia.
\end{itemize}

{\small
\bibliographystyle{ieee_fullname}
\bibliography{egbib}

\begin{thebibliography}{1}\itemsep=-1pt

\bibitem{cao2020delg}
Bingyi Cao, Andre Araujo, and Jack Sim.
\newblock Unifying deep local and global features for image search.
\newblock In {\em Proc. ECCV}, 2020.

\bibitem{DataCommons}
Data Commons.
\newblock Data commons place explorer.
\newblock \url{https://datacommons.org/place/}.
\newblock Accessed on February, 2021.

\bibitem{Jo_2020}
Eun~Seo Jo and Timnit Gebru.
\newblock Lessons from archives: Strategies for collecting sociocultural data
  in machine learning.
\newblock {\em Proceedings of the 2020 Conference on Fairness, Accountability,
  and Transparency}, Jan 2020.

\bibitem{mehrabi2019survey}
Ninareh Mehrabi, Fred Morstatter, Nripsuta Saxena, Kristina Lerman, and Aram
  Galstyan.
\newblock A survey on bias and fairness in machine learning, 2019.

\bibitem{Neyman_1992_Stratified_Sampling}
Jerzy Neyman.
\newblock On the two different aspects of the representative method: the method
  of stratified sampling and the method of purposive selection.
\newblock In S. Kotz and N.~L. Johnson, editors, {\em Breakthroughs in
  Statistics}. Springer, 1992.

\bibitem{Philbin2007OxfordDataset}
James Philbin, Ondrej Chum, Michael Isard, Josef Sivic, and Andrew Zisserman.
\newblock Object retrieval with large vocabularies and fast spatial matching.
\newblock In {\em 2007 IEEE Conference on Computer Vision and Pattern
  Recognition}, pages 1--8, 2007.

\bibitem{Philbin2008ParisDataset}
James Philbin, Ondrej Chum, Michael Isard, Josef Sivic, and Andrew Zisserman.
\newblock Lost in quantization: Improving particular object retrieval in large
  scale image databases.
\newblock In {\em 2008 IEEE Conference on Computer Vision and Pattern
  Recognition}, pages 1--8, 2008.

\bibitem{PopulationPyramid}
PopulationPyramid.net.
\newblock Population of {WORLD} 2019.
\newblock \url{https://www.populationpyramid.net/}.
\newblock Accessed on February, 2021.

\bibitem{weyand2020gldv2}
Tobias Weyand, André Araujo, Bingyi Cao, and Jack Sim.
\newblock Google landmarks dataset v2 - a large-scale benchmark for
  instance-level recognition and retrieval.
\newblock In {\em Proc. IEEE Conference on Computer Vision and Pattern
  Recognition (CVPR)}, 2020.

\end{thebibliography}
}

\end{document}